\documentclass{article}

    \PassOptionsToPackage{numbers, compress}{natbib}

\usepackage{enumitem}
\usepackage{amsmath} 
 \usepackage[preprint]{neurips_2026}

\usepackage[utf8]{inputenc} 
\usepackage[T1]{fontenc}    
\usepackage{hyperref}       
\usepackage{url}            
\usepackage{booktabs}       
\usepackage{amsfonts}       
\usepackage{nicefrac}       
\usepackage{microtype}      
\usepackage{xcolor}         
\usepackage{wrapfig}
\usepackage{graphicx}
\usepackage{multirow}
\usepackage{tabularray}
\usepackage{subcaption}
\usepackage{longtable}
\usepackage{tcolorbox}

\title{Deployment-Centered Evaluation: Predicting Query-Level Rejection Risk in a Clinical LLM System}

\author{Alyssa Unell \\
  Department of Computer Science\\
  Stanford University\\
  Stanford, CA 95305 \\
  \texttt{aunell@stanford.edu} \\
  \And
  Miguel Fuentes \\
  Department of Medicine\\
  Stanford University\\
  Stanford, CA 95305  \\
  \And
  Brenna Li \\
  Department of Medicine\\
  Stanford University\\
  Stanford, CA 95305 \\
  \texttt{brennali@stanford.edu} \\
    \And
  Bridget Lin \\
  Department of Biomedical Data Science\\
  Stanford University\\
  Stanford, CA 95305 \\
  \texttt{bridget7@stanford.edu} \\
  \And
  Meena Jagadeesan \\
Department of Computer Science\\
  Stanford University\\
  Stanford, CA 95305 \\
  \texttt{meenaj@seas.upenn.edu} \\
  \And
  Sanmi Koyejo \\
    Department of Computer Science\\
  Stanford University\\
  Stanford, CA 95305 \\
  \texttt{sanmi@stanford.edu} \\
    \And
  Nigam H. Shah \\
  Department of Medicine\\
  Stanford University \\
  Stanford, CA 95305  \\
  \texttt{nigam@stanford.edu} 
}

\begin{document}

\maketitle

\begin{abstract}
  Large language models (LLMs) are increasingly integrated into clinical systems, making it essential to evaluate the real-world utility of these systems. However, static benchmarks tend to measure correctness rather than user acceptance, aggregate performance across queries, and require densely annotated datasets---leading to major blind spots for evaluating clinical systems. In this work, we perform a deployment-centered evaluation of an LLM system embedded within electronic health records at an academic medical center, where user feedback is sparse but closely reflects the deployment conditions. Specifically, we train a pre-response classifier that estimates the risk that a future interaction will result in the user rejecting the LLM response, based on query content and deployment-specific context available before generation. We conduct a prospective analysis of our model over 4.5 months of user feedback, finding that our prediction model achieves an AUROC of 0.719. Further, we estimate the benefit of such predictions in two downstream use cases (guardrail triggering and abstention). Our key conceptual insight is that making use of deployment-specific context (i.e., the provider type, department name, language model used for response), as opposed to only query content, improves the ability to predict whether the user will reject the system output. Altogether, our empirical case study demonstrates the feasibility of predicting user rejection using deployment-specific context, opening the door to targeted guardrails. 
\end{abstract}

\section{Introduction}
As large language models (LLMs) become more capable, these models are increasingly deployed in clinical settings \cite{Singhal2023, griot2025deploy, chua2024deploy}. When users collaborate with these models, new opportunities emerge to improve both the efficiency and quality of care. However, this paradigm of user-LLM collaboration introduces a family of challenges that must be addressed: models can produce errors such as hallucinations or outputs that fail to meet the needs of their intended users \cite{Asgari2025framework}. When such failures go undetected, they impose additional verification burden on users \cite{Massenon2025My} or can foster over-reliance on the collaborating model \cite{Grolleau2026}.

While benchmarks make important progress in evaluating models on relevant tasks, these static evaluations have three main blind spots. First, most benchmark-centered evaluation prioritizes correctness as a target metric \cite{wang2025novelevaluationbenchmarkmedical, Bedi2026, Jiang2025}. Second, they capture aggregate performance across historical queries rather than predicting query-level performance on incoming ones. Third, they rely on curated, densely annotated datasets that are expensive to construct and difficult to refresh.

Deployment-centered evaluation requires a different lens. Rather than correctness alone, what matters is whether users would "reject" an output. Rejection is a noisy construct that approximately reflects a low perceived utility. Rejection may occur even when the output is objectively correct because it failed to meet the user's stylistic or contextual expectations. Instead of aggregate retrospection, we need forward-looking, query-level predictions that enable immediate action: invoking guard rails, deploying additional evaluation safeguards \cite{chung2025verifactverifyingfactsllmgenerated, oukelmoun2025detecting, care2026}, or abstaining. Such signals would ideally be derived before seeing the model's response to reduce prediction latency and decrease model usage costs. Rather than densely curated annotations, we work with what real-world deployments actually provide: sparse user feedback.

\begin{figure}[t]
\centering
\includegraphics[width=.9\linewidth]{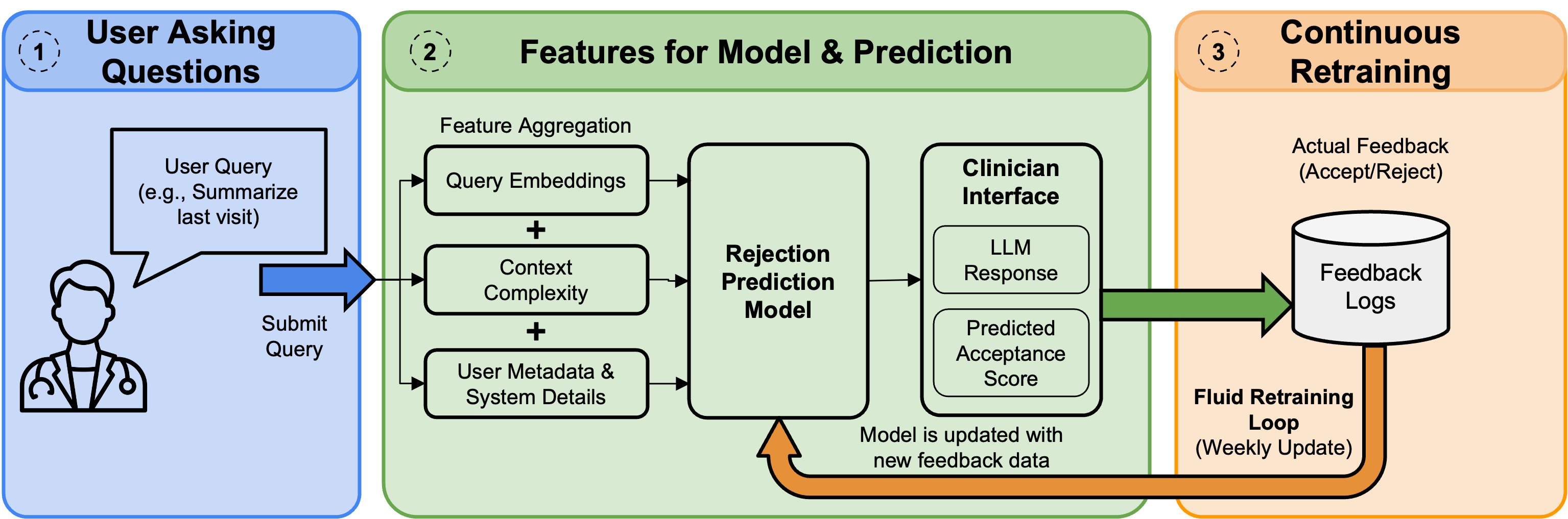}
\caption{\textbf{Overview of clinically deployed system.} Simplified system view showing the progression from \textbf{1} user input, through \textbf{2} feature-based prediction, to \textbf{3} the continuous retraining loop based on real-world feedback.}
\label{fig:fig1}
\end{figure}

In this work, we perform a deployment-centered evaluation on an LLM-based system embedded within electronic health records in an academic medical center. We train a pre-response model to predict rejection on future queries, leveraging the historical feedback provided by users of the system as training data. Figure~\ref{fig:fig1} illustrates the end-to-end architecture of our model, which is embedded into the clinical system. Our contribution is an empirical case study which shows the feasibility of deployment-centered evaluation, highlighting the role of deployment-specific context for rejection prediction.\footnote{We do not release an associated benchmark due to PHI-constraints.} 


We summarize our contributions below: 
\begin{itemize}[leftmargin=*]

\item \textbf{Prospective Evaluation:} We train a pre-response rejection model using the feedback provided by users on past queries. We conduct a prospective analysis using 4.5 months of user feedback, finding that the model achieves an AUROC of 0.719 on the task of predicting user rejection. We achieve 0.591 macro F1 and 0.651 micro F1 with consistent weekly performance, outperforming baselines and highlighting the role of deployment-specific context for improved rejection prediction.

\item \textbf{Downstream Use:} We frame rejection prediction as a deployable decision problem, demonstrating that our model can serve both abstention (high precision, $\beta=0.12$: 0.88 precision) and guardrail activation (high recall, $\beta=4.0$: 0.99 recall) by adjusting the operating threshold, without retraining.

\item \textbf{Role of Metadata}: Our key conceptual insight is that leveraging deployment-specific context (i.e., provider type, department name, model) improves performance, resulting in a 16.3\% improvement in AUROC compared to classifiers trained with query embeddings alone. The intuition is that user rejection criteria vary across departments and provider types, as we illustrate via a qualitative analysis. 
    
\end{itemize}

Ultimately, these results establish the feasibility of predicting user rejection in LLM-based clinical systems. Our pre-response rejection model overcomes key blind spots of benchmark-centered evaluation, enabling downstream use-cases such as abstention and guardrail activation.   
At a conceptual level, we leverage deployment-specific context to offset data sparsity challenges. More broadly, our work serves as a starting point for deployment-specific evaluations that move from benchmarks to interventions.  


\section{Related Work}

\subsection{LLM Evaluation for Clinical Tasks}
Traditionally, the paradigm for evaluating LLMs in clinical settings has relied on medical licensing examinations and multiple-choice benchmarks, obscuring critical failures in clinical reasoning while prioritizing correctness as the key measurement construct \cite{kung2023performance, hager2024evaluation}. A 2025 systematic review of 519 studies found that only 5\% utilized real patient care data for evaluation, highlighting a gap between clinical evaluation and practical implementations~\cite{bedi2025review}. Further studies reveal that models which appear competent on static exams often fail at proactive history-taking, iterative reasoning, and adherence to established diagnostic guidelines~\cite{johri2025evaluation, hager2024evaluation}.

While comprehensive benchmarks like MedHELM~\cite{Bedi2026} and MedAgentBench~\cite{Jiang2025} have expanded the task surface to include EHR interactions and complex tool-use, they continue to report aggregate performance based on objective ``correctness". Our work extends this literature by shifting the evaluation target from correctness to clinician ``acceptance". By leveraging explicit feedback from a live-deployed system, we demonstrate that a model's utility is defined not just by its accuracy, but also by its ability to meet the contextual needs of the individual user.

\subsection{Deployment-Centered Evaluations}
Evaluating real-world deployments poses a unique set of challenges. Benchmark performance is often poorly correlated with an application's real-world performance~\cite{liao2021we}, an issue that is driving a shift toward open-world evaluations: small-sample, often qualitative analyses of long-horizon deployments that aim to surface the failure modes that aggregate benchmark scores can both underestimate and overestimate~\cite{kapooropen}. Even when models perform well on held-out test sets, deployment success depends on whether users find the output useful enough to be incorporated into their workflow~\cite{he2019practical}. This gap has consequences: despite hundreds of regulator-approved machine learning tools internationally, large-scale uptake into routine clinical practice has proved elusive~\cite{scott2024achieving}. A finding from the health informatics literature suggests that digital health technologies can fail entirely when users abandon the system due to lack of trust, relevancy, or actionability~\cite{artsi2025large}. Regardless of objective accuracy, users’ acceptance of an output therefore has a lot of weight, yet clinicians’ acceptance of digital technologies is not uniform across the clinical workforce~\cite{stevens2023theory}. Studies of clinical decision support systems have further shown that acceptance varies by provider type, clinical specialty, and depth of integration with existing workflows~\cite{kilsdonk2017factors} — patterns consistent with our empirical finding that metadata encoding these contextual factors substantially improves acceptance anticipation. Our work addresses this gap directly: rather than evaluating an LLM on curated benchmarks, we perform a prospective analysis over 4.5 months of live user interactions within an EHR-embedded system, treating per-query acceptance — as signaled by clinician feedback — as the target metric.

\subsection{AI Measurement}

The majority of existing evaluations of AI systems anchor on correctness as the primary measurement target. Benchmarks such as HELM, MMLU, and BIG-Bench estimate and rank model capability using metrics such as accuracy, exact match (EM), and reference-based generation measures such as ROUGE and BLEU scores \cite{liang2023helm, mmlu2021, bigbench2023}. However, this induces a construct mismatch in real-world settings, where the utility of a response depends not only on correctness, but on whether it satisfies the needs, expectations, and constraints of the end user.
Recent work has sought to address this limitation by incorporating human judgment. For example, ChatBotArena uses an Elo-style rating system to rank models based on pairwise human judgments \cite{mtbench2023, chatbotarena2024}. However, this framework remains fundamentally comparative and aggregate, providing global model rankings rather than measuring per-query utility. As a result, these approaches answer a different evaluation question, and do not assess whether a single response will be accepted in a given user and deployment context. A complementary body of work aims to predict query-level performance. Measurement frameworks such as item Response Theory (IRT) offer a more principled measurement framework by modeling the probability of a correct response as a function of latent model ability and item difficulty \cite{lord1968statistical, lord1980applications}. This approach, and higher-dimensional generalizations, provide a more granular item-level view of model capability across varying task complexities \cite{zhou2026, polo2024, truongIRT}. However, such approaches still target correctness, which may diverge from operational acceptance and fail to capture user-specific variability. 
\section{Methods}

We formalize our task of predicting query-level rejection (Section \ref{subsec:task}), outline our training approach for the pre-response rejection model (Section \ref{subsec:methodtraining}), and describe our dataset processing (Section \ref{subsec:datasetprocessing}). 

\subsection{Task Formalization}\label{subsec:task}

\paragraph{Deployed LLM system.} We study a deployed LLM system integrated directly into the electronic health record (EHR) environment to minimize workflow friction. The system operates within an academic medical center and routes user queries to one of two PHI-compliant language models hosted via Azure OpenAI \cite{MicrosoftAzureOpenAI2026}. Following each generated response, the user interface presents a feedback instrument consisting of "thumbs up" and "thumbs down" icons. Providers can optionally click these icons to submit explicit binary feedback, alongside optional free-text commentary. We define a "thumbs up" as an \textbf{acceptance} of the output and a "thumbs down" as a \textbf{rejection}. Deployment-specific context such as user metadata and system information accompany each query. Figure~\ref{fig:fig1} illustrates the deployment ecosystem and our prediction model. This evaluation was reviewed by the academic medical center's Institutional Review Board (IRB Protocol \#86364, NHSR determination) and classified as non-human subject research.

\paragraph{Data representation.}
We represent the data as follows. Let $Q = \{q_i\}_{i=1}^N$ the set of user-submitted queries. For each query $q_i \in Q$, the LLM system generates the corresponding response. Each query $q_i$ is also associated with the following deployment-specific context. This includes \textit{user context}, which consists of the provider type $p_i$ and department name $d_i$, and \textit{system context}, which consists of the model type $m_i$ of the specific LLM to which the query was routed and the length $l_i$ of the associated patient record used in the payload. 

We let $b_i = 1$ if the user left binary structured feedback (accept vs. reject) on the query, and $b_i = 0$ otherwise. We represent structured feedback for query $q_i$ as $r_i \in \{-1, 1\}$, where:
\begin{itemize}[leftmargin=*]
    \item $r_i = 1$ indicates rejection, and
    \item $r_i = -1$ indicates acceptance.
\end{itemize}

\paragraph{Our goal.}
Our goal is to learn a score function $f(\cdot)$ that predicts query-level rejection $r_i$ from the query $q_i$, and the deployment-specific context including the user metadata $(p_i, d_i)$ and system information $(m_i, l_i)$. The score $f(q_i, p_i, d_i, m_i, l_i)$ represents the predicted acceptance probability. We evaluate the score function $f$ on the basis of AUROC and F1.

\subsection{Pre-Response Rejection Prediction Model}\label{subsec:methodtraining}

When we train and evaluate our model, we restrict to the subset
$Q^{*} = \{ q_i \in Q \mid b_i = 1 \}$, which captures queries with explicit binary feedback.


\paragraph{Logistic regression approach.} 
To learn the score function $f$, we train a logistic regression model trained with 3000 maximum iterations and balanced class weighting. Model training and evaluation were executed within a HIPAA-compliant Databricks environment utilizing a single CPU node, reflecting the lightweight compute requirements of our approach. This model includes a subset of the variables $(q_i, p_i, d_i, m_i, l_i)$ as features. Note that we represent the query $q_i$ in terms of the embedding computed using text-embedding-3-large from OpenAI \cite{OpenAI2024embeddings}. We let the provider type $p_i$, department name $d_i$, and  model type $m_i$
denote one-hot encodings, which have dimensionality 17, 129, and 2, 
respectively.  

\paragraph{Choosing a feature set.}
A key design problem is to determine which subset of variables to include in order to maximize AUROC. We consider all possible feature set combinations with the exemption of the empty set, resulting in 31 possible combinations. We train all 31 possible models on the initial train set, and we  select the best model on the validation set  (see Section \ref{subsec:datasetprocessing} for a description of the initial train set and validation set). Specifically, we perform a grid search over logistic regression hyperparameters using the SAGA solver with $L_2$ and elastic net penalties. We vary the inverse regularization strength $C\in{0.001, 0.005, 0.01, 0.05, 0.1, 0.5, 1}$ and the elastic net mixing parameter $l1_{ratio}\in{0.05, 0.5, 0.7, 0.9, 1.0}$; pure L2 models are evaluated over the same C grid. This gives us our prediction model. We dynamically retrain this model by including new data as it arrives, as we discuss in Section \ref{subsec:datasetprocessing}. 

\paragraph{Baselines.}
We consider several baselines. The first baseline is to train a logistic regression model with the feature set consisting only of the user query embeddings $q_i$. Another set of baselines uses LLM-judges to return the probability of acceptance for a given interaction. We consider two LLM-judge formulations: one with user query only and one with user query plus deployment-specific context information that matches our best model's selected feature set. We use GPT-4.1 with default temperature 1.0 as the LLM-judge \cite{gpt4} with the associated prompt in Appendix~\ref{app:prompt}. For our analysis of downstream use-cases which require us to select classification thresholds, we consider a few additional baselines: an always accept classifier, an always reject classifier, and LLM-judge classifiers as described above that return discrete accept or reject labels instead of continuous scores.

\subsection{Dataset Processing}\label{subsec:datasetprocessing}

\paragraph{Data filtering.}
We collect 74,729 total interactions from logs between 09/01/2025 and 04/11/2026. 1,196 of these interactions have explicit binary feedback. In the case of multi-turn interactions, we remove all but the first query posed by the user. This filtering leaves 985 unique interactions. 80 entries are missing provider type and 16 entries are jointly missing model type and context length. This leaves 889 interactions that are uniquely annotated and contain all system- and user-level data. Finally, 11 queries are used in system red-teaming as denoted by a pre-determined red-teaming comment being present in the open-ended user feedback. This leaves us with a final annotated dataset of 878 total query-annotation pairs. 

We compare the distribution of annotated and non-annotated queries in Appendix~\ref{app:distribution}, highlighting that while the distributions are roughly aligned, there is over and under representation of specific departments and providers with respect to feedback rate. Specifically, residents are frequent users who do not provide feedback while registered nurses have a much higher base feedback rate. Similarly, primary care departments and inpatient units have higher base feedback rates, as well. We defer a careful analysis of the impact of this selection bias to future work. 

\paragraph{Initial train set.}
We build the initial train set as follows. Our initial training data consists of the 283 queries from the initial 8 weeks of annotated user interactions (09/01/2025 to 10/30/2025). We drop queries that are duplicates of query and user ID asked within the same 1 hour time period. 1 query is dropped in our initial training data.

\paragraph{Validation set.}
We build the validation set from the 4-week period (10/30/2025 to 11/27/2025), which  encompasses 162 annotated queries. Following the initial train set deduplication, we drop queries that are duplicates of query and user ID asked within the same 1 hour time period. 5 queries are dropped in our validation data. We select the feature set and hyperparameters that maximize AUROC on the validation set, freezing for all downstream analysis. In our downstream use-case analysis, we also use this validation set to select an acceptance/rejection threshold.

\paragraph{Prospective analysis.}
We conduct a prospective analysis over 19 weeks of test data (11/27/2025 to 04/11/2026), measuring the rejection likelihood of each query on a weekly basis.  After conducting a prospective analysis for a given week, we dynamically incorporate that week of data with the true user labels into the training data of our model and retrain the model from scratch using all of the data collected thus far, keeping the feature set and hyperparameters frozen. This represents a rolling train window that allows us to benefit from additional usage patterns, continuously adapting the classification model to user-LLM trends. We de-duplicate any repeats within the same hour. 13 additional queries are dropped during training over the following 19 weeks due to this filtering step.
\section{Results}

We report the performance of our prospective analysis in Section \ref{subsec:performance}, discuss downstream use-cases in Section \ref{subsubsec:abstention}, and analyze the role of deployment-specific context in Section \ref{subsec:metadata}.

\subsection{Prospective Analysis}
\label{subsec:performance} 
We evaluate our model along two complementary metrics: AUROC and F1. 

\paragraph{AUROC.} AUROC captures the full tradeoff between sensitivity and specificity across all possible decision thresholds, without requiring practitioners to commit to a single operating point. This is particularly important in clinical settings, where individual risk tolerance and flagging preferences vary: a high-specificity operating point may be preferred in some contexts, while a high-sensitivity one in others. The ROC curve characterize model behavior across potential downstream use-cases.

\begin{figure}[!htbp]
\centering
\begin{subfigure}[c]{0.5\linewidth}
    \centering
    \includegraphics[width=\linewidth]{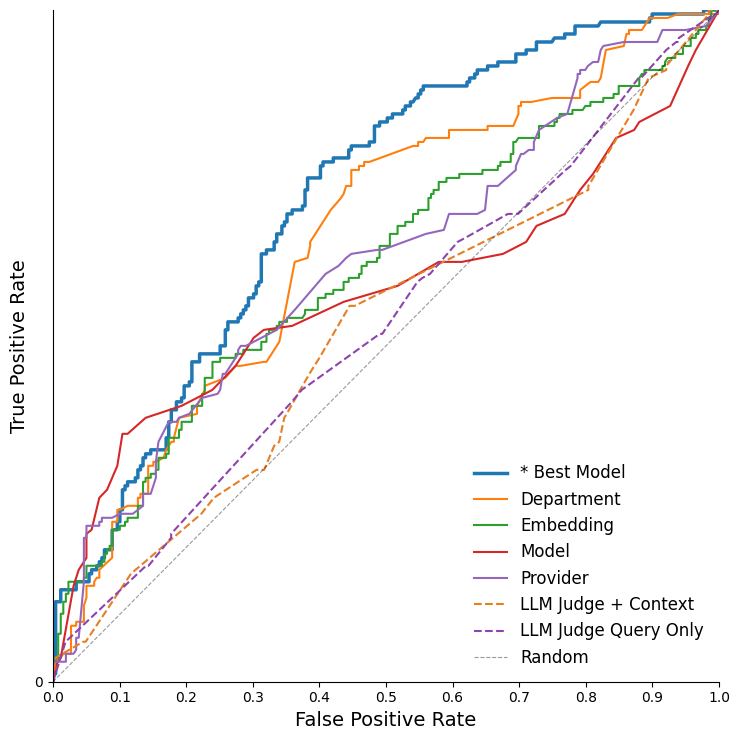}
\end{subfigure}
\hfill
\begin{subtable}[c]{0.45\linewidth}
    \centering
    \begin{tabular}{lc}
    \hline
    \textbf{Predictor} & \textbf{AUROC [95\% CI]} \\
    \hline
    \textbf{Best Model}          & \textbf{0.719 [0.670--0.767]} \\
    Department          & 0.661 [0.605--0.715] \\
    Provider            & 0.625 [0.569--0.678] \\
    Embedding           & 0.618 [0.562--0.675] \\
    Model               & 0.578 [0.518--0.639] \\
    LLM Judge Query Only & 0.530 [0.475--0.586] \\
    LLM Judge + Context & 0.521 [0.464--0.577] \\
    Random              & 0.500 [0.500--0.500]\\
    \hline
    \end{tabular}
\end{subtable}
\caption{\textbf{Our model results in 0.719 AUROC.}
Models are trained on historical data, validated for model selection, and
evaluated on held-out future data.}
\label{fig:fig2}
\end{figure}

Focusing on AUROC, we perform a prospective analysis on our model as described in the previous section. We select a \textit{Best Model} from the AUROC on the validation set. This \textit{Best Model} has the following feature set: \textbf{Embedding + Provider + Department + Model}. This \textit{Best Model} achieves an AUROC of 0.719 across test windows with a confidence interval of [0.670, 0.767] obtained via bootstrapping with 2000 iterations. This illustrates that rejection can be feasibly predicted using the available data. Figure \ref{fig:fig2} shows the AUROC of all test data over 19 weeks of user interactions. 

Our model achieves a higher AUROC than several baselines, including a model trained on query embeddings alone (i.e., the feature set $\left\{q_i\right\}$) and LLM Judge classifiers with and without the same deployment-specific context feature set as our \textit{Best Model}. Always accept/Always reject baselines are omitted here for redundancy, as these approaches would both have a prospective AUROC of 0.500.

\begin{figure*}[!htbp]
\centering

\begin{minipage}[c]{0.52\textwidth}
    \centering
    \includegraphics[width=\textwidth]{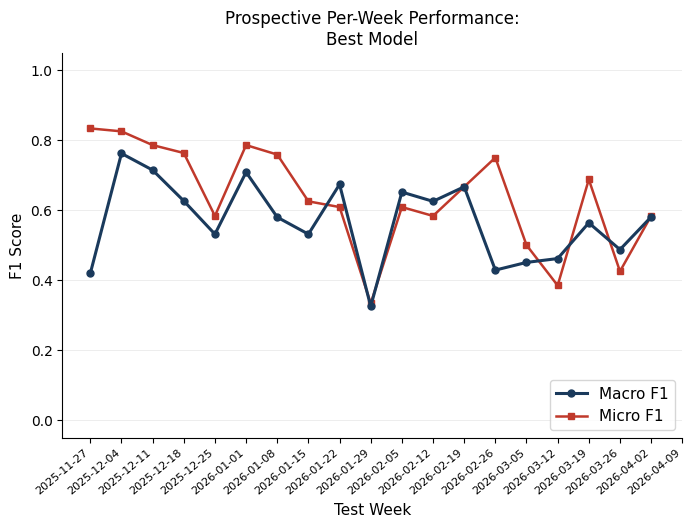}
\end{minipage}
\hfill
\begin{minipage}[c]{0.45\textwidth}
    \centering
    \small
    \begin{tabular}{lrr}
    \toprule
    \textbf{Method} & \textbf{Macro F1} & \textbf{Micro F1} \\
    \midrule
    Best model           & \textbf{0.591} & \textbf{0.651} \\
    Emb-only             & 0.556          & 0.637          \\
    LLM Judge + Context & 0.488          & 0.553          \\
    LLM Judge Query Only & 0.478          & 0.499          \\
    Always Reject        & 0.269          & 0.393          \\
    Never Reject         & 0.368          & 0.607          \\
    \bottomrule
    \end{tabular}
\end{minipage}

\caption{\textbf{Model performance over weekly deployment.} \textit{(Left)} Per-week
Macro and Micro F1 of each method across the test period, demonstrating robustness to
distribution shift over deployment. \textit{(Right)} Weighted average F1 across test
weeks (each week weighted by sample count), evaluated at independently tuned thresholds. Bold indicates the best value per column.}
\label{fig:f1_results}
\end{figure*}

\paragraph{F1.} Since practitioners will ultimately use the model to make classification decisions (e.g., abstention or guardrail activation), we evaluate decision quality against baselines using two F1 variants: Macro F1 (unweighted mean across classes, sensitive to minority-class performance) and Micro F1 (globally aggregated, reflecting overall accuracy under class imbalance), as reported in  Table \ref{fig:f1_results}. 

To make classification decisions, it is necessary to select a classification threshold. To simulate such deployment, we select the threshold to maximize the reported F1 variant on the validation set. We evaluate the F1 scores of our best model and compare against a suite of baselines, including models trained on query embeddings alone, LLM-judge baselines, and naive always/never reject baselines as shown in Table \ref{fig:f1_results}. Specifically, this shows that our model achieves the highest macro and micro F1 scores when compared to baseline approaches. Additionally, Figure \ref{fig:f1_results} illustrates that our best model shows minimal performance degradation across weeks, validating the temporal stability of the identified features for rejection prediction.

\subsection{Downstream Use-Cases}
\label{subsubsec:abstention}

We next explore how to leverage our model to support two downstream
use-cases: \textit{abstention} and \textit{guardrail activation}. These use-cases motivate different operating thresholds, which we parameterize via $F_\beta$, as defined in Appendix~\ref{app:fbeta}. A smaller $\beta$ weights precision more heavily, appropriate when false positives are costly (e.g., unnecessarily withholding a useful response); a larger $\beta$ weights recall, appropriate
when missing a true rejection carries greater risk (e.g., surfacing a low-quality response in a high-stakes setting). For each $\beta$, we tune the decision threshold based on the value of $F_\beta$ on the
validation set. We report test-set performance in Table~\ref{tab:fbeta_results}, reflecting
realistic prospective deployment performance.

\paragraph{Abstention}
At low $\beta$ (e.g., $\beta=0.12$, $\beta=0.25$), the model operates in a
high-precision regime suitable for \textit{abstention}: preemptively filtering queries the
system is likely to reject, sparing users from low-quality responses and reducing
verification fatigue~\cite{ohde2025verification-burden, machcha2026abstention}. At $\beta=0.12$, precision reaches 0.88 with only
3 false positives across the entire test set, making abstention feasible without
meaningfully disrupting users who would have received a successful response.

\paragraph{Guardrail Activation}
At higher $\beta$ (e.g., $\beta=2.0$, $\beta=4.0$), the model shifts toward high recall,
recovering 94--99\% of true rejections. This regime is better suited to softer
interventions such as surfacing a warning, triggering additional verification, or invoking
hallucination-detection pipelines~\cite{chung2025verifactverifyingfactsllmgenerated,
Asgari2025framework}. Here the cost of a false positive is lower---the user sees an
unnecessary flag rather than losing access to a response entirely---so the higher false
positive rate is acceptable in exchange for near-complete recall of problematic queries.


\begin{table}[t]
\centering
\small
\begin{tabular}{rrrrrrrrr}
\toprule
$\beta$ & Threshold & TP & FP & TN & FN & Precision & Recall & $F_\beta$ \\
\midrule
0.12 & 0.833 & 22 & 3 & 256 & 146 & 0.88 & 0.13 & 0.809 \\
0.50 & 0.565 & 80 & 55 & 204 & 88 & 0.59 & 0.48 & 0.565 \\
1.00 & 0.365 & 138 & 125 & 134 & 30 & 0.52 & 0.82 & 0.640 \\
2.00 & 0.206 & 158 & 188 & 71 & 10 & 0.46 & 0.94 & 0.776 \\
4.00 & 0.102 & 167 & 248 & 11 & 1 & 0.40 & 0.99 & 0.915 \\
\bottomrule
\end{tabular}
\caption{Embedding + Provider + Department + Model performance at each F$_\beta$ operating point. TP = true rejects, FP = false rejects, TN = true accepts, FN = missed rejects. Threshold tuned per $\beta$ on the validation set; stats evaluated on the test set. Duplicate operating points (identical threshold across $\beta$ values) are shown once. Higher $\beta$ weights recall more heavily; lower $\beta$ weights precision.}
\label{tab:fbeta_results}
\end{table}

\subsection{Role of Deployment-Specific Context in Rejection Prediction}
\label{subsec:metadata}

The best performing model includes query embeddings, provider type $p_i$, department name $d_i$, and model $m_i$. Notably, this feature set includes 
both user and system context, suggesting that deployment-specific context improves prediction power in this data sparse setting. In this section, we investigate this hypothesis quantitatively and qualitatively. 

\subsubsection{Quantitative Perspective}

Our \textit{Best Model} achieves a 16.3\% relative improvement over the query embedding-only baseline. This highlights the value of leveraging department-specific context. In fact, removing embedding from the feature set (but retaining provider type, department name, and model name) achieves an AUROC that is only slightly lower than the \textit{Best Model} (Appendix~\ref{app:allfeatures}). This suggests that deployment-specific context may be even more informative than the embedding, especially in the data sparse settings in which our evaluation operates.

\begin{wrapfigure}{r}{0.32\textwidth}
\vspace{-1.2em}
\centering
\small
\begin{tabular}{lr}
\toprule
Feature & Count (of 10) \\
\midrule
Department      & 9 \\ 
Model       & 8 \\ 
Embedding     & 7 \\ 
Provider     & 6 \\
Context length &  3\\
\bottomrule
\end{tabular}
\caption{Feature occurrence counts across the top 10 configurations ranked by
validation AUROC.}
\label{tab:top20}
\vspace{-1em}
\end{wrapfigure}

We also examine the impact of individual features through two different lenses. First, we report AUROC for the top 10 performing feature sets, quantifying the frequency of each feature in this subset (Table~\ref{tab:top20}). This coarse analysis results in the following order of feature importance: department, model, embedding, provider, and then context length. Second, we consider the AUROC of a model trained on each feature in isolation (Figure~\ref{fig:fig2}; Appendix~\ref{app:allfeatures}). This alternative analysis results in the following feature ordering: department, provider, embedding, model, and then context length. Both of these analyses highlight the importance of department name and provider type.  






\subsubsection{Qualitative Perspective}\label{subsec:qualitative}

To better understand why department name and provider type improve predictive accuracy, we perform a qualitative analysis on the feedback provided by the users in our system. We observe significant variation in rejection probabilities across provider types and department names, as we illustrate through rejection statistics and specific examples from user feedback.  We report the raw rejection rate by the most frequently represented features in Appendix~\ref{subsec:combined_features}, illustrating that physicians have above average rejection rates, while nurses have below average rejection rates. Users from primary care accept more frequently compared to users from administration departments and oncology departments. 

\paragraph{Role of provider.}
To illustrate provider effects on rejection, Table~\ref{tab:qual_results}(a) presents representative prompts and acceptance labels for similar queries across provider types within the same department. This mirrors a broader pattern in our data: among physicians in the cardiology department, 9 of 10 responses were rejected, whereas among nurse practitioners only 1 of 40 was rejected (Appendix~\ref{tab:combined_features}). When rationale was provided, physicians cited inaccuracies and omissions of expected clinical detail, suggesting they apply a higher specificity threshold than nurse practitioners, who may find high-level summaries sufficient for their workflows. This pattern indicates that acceptance reflects not only objective response quality but also subjective, role-dependent expectations of what constitutes relevant and sufficient information.

\begin{table*}[!htbp]
\centering
\small
\begin{tabular}{llp{5cm}p{4.5cm}}
\toprule
\textbf{Provider} & \textbf{Rejected} & \textbf{Prompt} & \textbf{Rationale} \\
\midrule
\multicolumn{4}{l}{\textit{(a) Role of provider type — Cardiology department}} \\
\midrule
Physician          & Yes  & Please provide a summary timeline of this man's history & This is the worst example I've seen of false information being presented \\
Physician          & Yes  & Write a one liner about this patient, including relevant objective data for today and plan. One paragraph. & It did not include relevant details about his dual chamber pacemaker, prior stenting, or the fact that he was admitted for cardiogenic shock \\
Nurse Practitioner & No & Summarize relevant clinical updates since the patient's last visit in the heart failure clinic & — \\
Nurse Practitioner & No & Describe cardiovascular events over the past 6 months & — \\
\midrule
\multicolumn{4}{l}{\textit{(b) Role of department — Physicians only}} \\
\midrule
Critical Care & Yes  & Detail cardiology visits by date, descending & — \\
Hospital      & No & Summary cardiac history & — \\
Radiology     & No & What is known about the patient's coronary artery disease & — \\
\bottomrule
\end{tabular}
\caption{Qualitative examples illustrating the role of provider type \textit{(a)} and
department \textit{(b)} in clinician rejection. \textit{(a)} contrasts physician and
nurse practitioner rejection within the cardiology department; \textit{(b)} shows
physician rejection across departments. Rationale is recorded only for rejected queries;
``—'' indicates no rationale provided.}
\label{tab:qual_results}
\end{table*}

\paragraph{Role of department.} Similarly, Table~\ref{tab:qual_results}(b) shows representative examples from physicians across departments with comparable queries that yield different acceptance outcomes. The Critical Care physician rejected the response while Hospital and Radiology physicians accepted theirs. Without rationale, the driver of this difference (whether factual inaccuracy, omission, or individual preference ) cannot be determined. Nonetheless, the pattern suggests that departmental context shapes acceptance criteria in ways that go beyond query content alone.

\section{Discussion}

In this work, we perform a deployment-centered evaluation on an LLM-based system embedded within electronic health records in an academic medical center. We train a pre-response rejection model which achieves an AUROC of 0.719 on a prospective analysis spanning 4.5 months of user feedback. 
By adjusting operating thresholds, the same model supports both high-precision abstention (0.88 precision at $\beta=0.12$) and high-recall guardrail activation (0.99 recall at $\beta=4.0$), illustrating its practical value for deployment decisions. 
Our key conceptual insight is that deployment-specific context (including provider type, department, and language model) improves prediction performance, yielding a 16.3\% relative gain in AUROC over models trained on query embeddings alone. This finding underscores that who is asking, in what context, and for what purpose matters as much as what is being asked. 


This work has limitations that should be explored in future work. By nature of a real-world clinical deployment evaluation, our data is limited to passively collected binary feedback, which only 1.6\% of users provide explicitly. This sparsity is an intrinsic property of deployment data, which constrains its statistical power and limits generalization, particularly for higher-dimensional features such as embeddings, where more data would likely yield additional signal.  Furthermore, the users who choose to provide feedback may differ from those who do not, introducing selection bias that we cannot fully characterize from observational data alone. The distribution of annotated data is similar but not identical to the distribution of unannotated data (Appendix \ref{app:distribution}); the annotated subset may over-represent queries where clinicians had stronger reactions. Future work could address these biases through targeted feedback elicitation in lower-response subgroups, qualitative studies probing non-response, or methods that leverage unlabeled user queries. Finally, while this analysis is the first of its kind to identify the role of deployment-specific context for predicting query-level acceptance, our results are drawn from a single academic medical center, with its own patient population, clinician composition, EHR configuration, and practices around AI tooling. Future work could extend these methodological and qualitative explorations to a broader range of deployment contexts.

\section{Conclusion}
Our work provides a starting point for a deployment-centered evaluation framework that overcomes key blind spots of traditional benchmark-centered evaluation. First, rather than evaluating correctness, we predict user rejection to better capture perceived utility in real workflows. Second, instead of measuring aggregate performance on historical queries, we predict rejection on future queries, enabling query-specific actions. Third, we leverage sparse user feedback collected during routine clinical use rather than curated, densely annotated benchmarks. By addressing these blind spots, our framework supports deployment-time interventions and reveals how real-world usage patterns shape outcomes. We demonstrate that "success" in deployed settings depends as much on the context surrounding a question as on the question itself.

\section{Acknowledgments}
MJ acknowledges partial support from a Stanford AI Lab Postdoctoral fellowship. SK acknowledges
support by NSF 2046795 and 2205329, IES R305C240046, ARPA-H, the MacArthur Foundation,
Schmidt Sciences, and HAI. AU acknowledges support from the Stanford Graduate Fellowship and NSF GRFP grant number DGE-2146755. NS acknowledges support from the Data Science Program at Stanford Healthcare.

\bibliographystyle{plainnat}
\bibliography{bib}


\appendix
\section{Appendix}
\label{sec:app}

\subsection{LLM-as-Judge Prompt}
\label{app:prompt}

We employed an LLM-as-judge approach to generate pseudo-labels for user query acceptance. 
Two prompt variants were evaluated: one providing only the user query (\textit{Query Only}), 
and one augmenting the query with session metadata (\textit{Query + Metadata}).

\subsubsection*{Base Prompt (Both Variants)}

\begin{quote}
\ttfamily\small
You are evaluating the following user query to determine how likely a user would be to accept an LLM-generated response to this query.

User Query: ``\{user\_query\}''

\{context\_block\}

Please evaluate this query on two dimensions:

1. CONTINUOUS SCORE (0--1): How likely is the user to accept an LLM response to this query?

2. DISCRETE ACCEPTANCE (Yes/No): Will the user accept an LLM response to this query?
\begin{itemize}
    \item Yes = The user will accept an LLM response
    \item No = The user would NOT accept an LLM response
\end{itemize}

Respond in EXACTLY this format:\\
CONTINUOUS\_SCORE: [number between 0 and 1]\\
DISCRETE\_ACCEPTANCE: [Yes or No]\\
REASONING: [brief explanation]
\end{quote}

\subsubsection*{Deployment-Specific Context Block (Query + Context Variant Only)}

When session metadata was available, the following context block was appended after the user query:

\begin{quote}
\ttfamily\small
Additional Context:\\
- Model: \{model\}\\
- Provider Type: \{provider\_type\}\\
- Department Name: \{department\_name\}
\end{quote}

Fields with null, empty, or undefined values were excluded from the context block. 
The \textit{Query Only} variant omitted the context block entirely, receiving only the 
user query as input.



\subsection{Distributional Comparison of Annotated vs. Non-Annotated Interactions}
\label{app:distribution}





\begin{figure}[htbp]
    \centering
    \begin{subfigure}{0.48\textwidth}
        \centering
        \includegraphics[width=\textwidth]{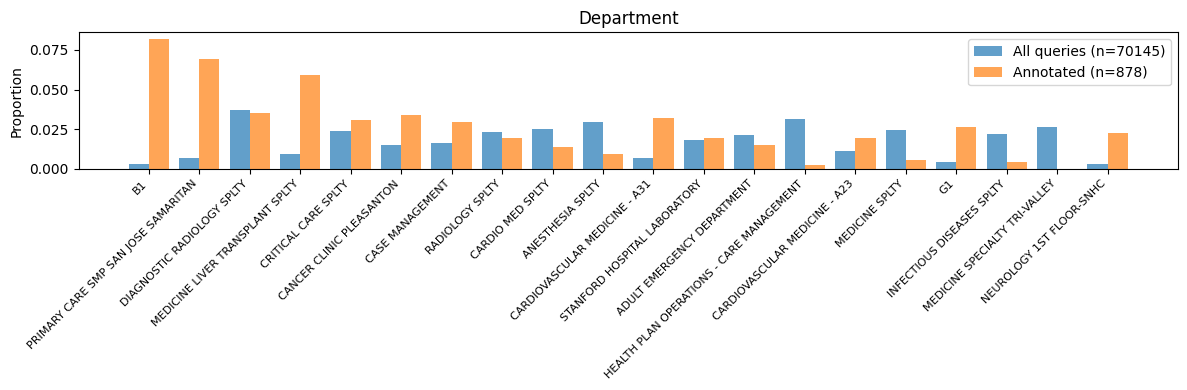}
        \caption{Department}
    \end{subfigure}
    \hfill
    \begin{subfigure}{0.48\textwidth}
        \centering
        \includegraphics[width=\textwidth]{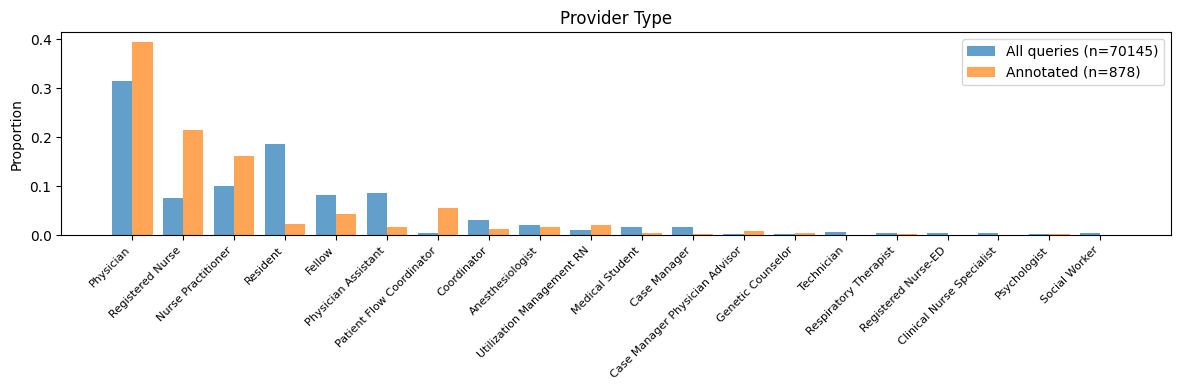}
        \caption{Provider Type}
    \end{subfigure}

    \vspace{0.75em}

    \begin{subfigure}{0.6\textwidth}
        \centering
        \includegraphics[width=\textwidth]{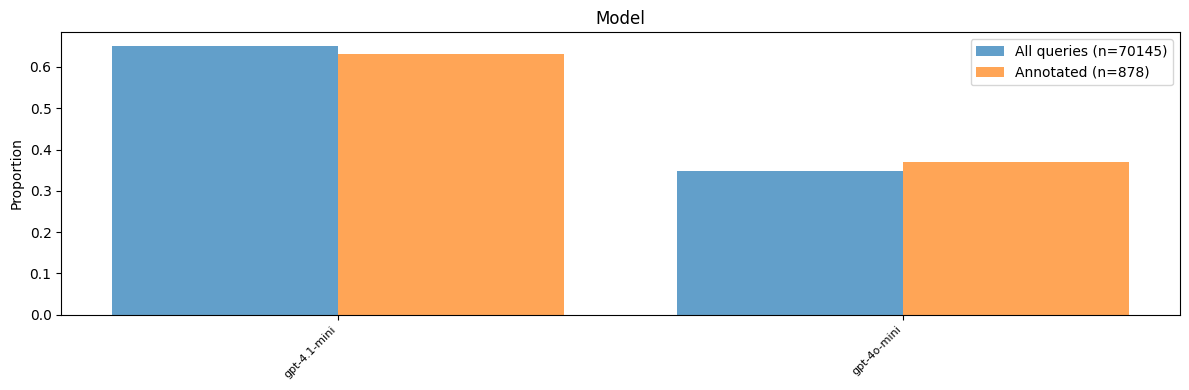}
        \caption{Model}
    \end{subfigure}

    \caption{Distribution comparison between all queries and annotated subset across key features.}
    \label{fig:dist_comparison}
\end{figure}

\subsection{F$_\beta$ Score at a Threshold}
\label{app:fbeta}
The F$_\beta$ score is a weighted harmonic mean of precision and recall, where $\beta > 0$ controls the relative importance of recall versus precision:
\begin{equation}
F_{\beta} = (1 + \beta^2) \cdot \frac{\mathrm{Precision} \cdot \mathrm{Recall}}{\beta^2 \cdot \mathrm{Precision} + \mathrm{Recall}}.
\end{equation}


\subsection{All Model Performances on Test Set.}
\label{app:allfeatures}
\begin{longtable}{lc}
\hline
\textbf{Model} & \textbf{AUROC (95\% CI)} \\
\hline
\endfirsthead
\hline
\textbf{Model} & \textbf{AUROC (95\% CI)} \\
\hline
\endhead
\hline
\endfoot
Embedding \texttt{+} Provider \texttt{+} Department \texttt{+} Model & 0.719 (0.670, 0.767) \\
Provider \texttt{+} Department \texttt{+} Model & 0.715 (0.664, 0.765) \\
Embedding \texttt{+} Department \texttt{+} Model & 0.708 (0.658, 0.758) \\
Embedding \texttt{+} Provider \texttt{+} Department \texttt{+} Model \texttt{+} Context\_length & 0.702 (0.651, 0.750) \\
Embedding \texttt{+} Department \texttt{+} Model \texttt{+} Context\_length & 0.689 (0.638, 0.741) \\
Embedding \texttt{+} Provider \texttt{+} Department & 0.689 (0.636, 0.739) \\
Embedding \texttt{+} Provider \texttt{+} Model & 0.685 (0.635, 0.735) \\
Department \texttt{+} Model & 0.684 (0.631, 0.737) \\
Provider \texttt{+} Department \texttt{+} Model \texttt{+} Context\_length & 0.683 (0.631, 0.734) \\
Embedding \texttt{+} Department & 0.681 (0.626, 0.730) \\
Embedding \texttt{+} Provider \texttt{+} Department \texttt{+} Context\_length & 0.679 (0.625, 0.731) \\
Provider \texttt{+} Department & 0.678 (0.622, 0.731) \\
Embedding \texttt{+} Department \texttt{+} Context\_length & 0.669 (0.616, 0.721) \\
Provider \texttt{+} Model & 0.668 (0.614, 0.720) \\
Embedding \texttt{+} Provider \texttt{+} Model \texttt{+} Context\_length & 0.667 (0.614, 0.718) \\
Embedding \texttt{+} Provider & 0.665 (0.611, 0.716) \\
Department \texttt{+} Model \texttt{+} Context\_length & 0.663 (0.609, 0.716) \\
Provider \texttt{+} Department \texttt{+} Context\_length & 0.663 (0.607, 0.715) \\
Department & 0.661 (0.605, 0.715) \\
Department \texttt{+} Context\_length & 0.651 (0.598, 0.704) \\
Embedding \texttt{+} Model & 0.651 (0.599, 0.705) \\
Embedding \texttt{+} Provider \texttt{+} Context\_length & 0.652 (0.597, 0.706) \\
Provider \texttt{+} Model \texttt{+} Context\_length & 0.640 (0.585, 0.693) \\
Provider & 0.625 (0.569, 0.678) \\
Embedding & 0.618 (0.562, 0.675) \\
Provider \texttt{+} Context\_length & 0.614 (0.558, 0.670) \\
Embedding \texttt{+} Context\_length & 0.601 (0.546, 0.656) \\
Embedding \texttt{+} Model \texttt{+} Context\_length & 0.629 (0.574, 0.685) \\
Model & 0.578 (0.518, 0.639) \\
Model \texttt{+} Context\_length & 0.573 (0.516, 0.632) \\
Context\_length & 0.471 (0.420, 0.526) \\
\hline
\caption{Prospective AUROC by model. CI computed via bootstrap (2000 resamples).}
\label{tab:prospective_auroc}
\end{longtable}

\newpage
\subsection{Raw Counts}
\label{subsec:combined_features}
\begin{table}[!htbp]
\centering
\begin{tabular}{l l c c}
\toprule
\textbf{Dataset} & \textbf{Feature} & \textbf{N\_total} & \textbf{Reject Rate} \\
\midrule
\multirow{1}{*}{Total Statistics} & ALL POINTS & 878 & 0.333 \\
\midrule
\multirow{2}{*}{Model} & gpt-4.1-mini & 554 & 0.255 \\
 & gpt-4o-mini & 324 & 0.466 \\
\midrule
\multirow{10}{*}{Department} & B1 & 72 & 0.250 \\
 & PRIMARY CARE SMP SAN JOSE SAMARITAN & 61 & 0.180 \\
 & MEDICINE LIVER TRANSPLANT SPLTY & 52 & 0.192 \\
 & DIAGNOSTIC RADIOLOGY SPLTY & 31 & 0.129 \\
 & CANCER CLINIC PLEASANTON & 30 & 0.567 \\
 & CARDIOVASCULAR MEDICINE - A31 & 28 & 0.036 \\
 & CRITICAL CARE SPLTY & 27 & 0.185 \\
 & CASE MANAGEMENT & 26 & 0.654 \\
 & G1 & 23 & 0.391 \\
 & NEUROLOGY 1ST FLOOR-SNHC & 20 & 0.050 \\
\midrule
\multirow{10}{*}{Provider Type} & Physician & 346 & 0.329 \\
 & Registered Nurse & 189 & 0.317 \\
 & Nurse Practitioner & 142 & 0.169 \\
 & Patient Flow Coordinator & 48 & 0.688 \\
 & Fellow & 39 & 0.256 \\
 & Resident & 21 & 0.381 \\
 & Utilization Management RN & 18 & 0.611 \\
 & Anesthesiologist & 14 & 0.429 \\
 & Physician Assistant & 14 & 0.357 \\
 & Coordinator & 12 & 0.583 \\
\bottomrule
\end{tabular}
\caption{Raw counts and rejection rate for all 878 user queries used in train, validation, and testing.}
\label{tab:combined_features}
\end{table}

\end{document}